\documentclass{article} 
\usepackage{iclr2026_conference,times}
\usepackage{microtype}
\usepackage{graphicx}
\usepackage{subcaption}
\usepackage{booktabs} 
\usepackage[HTML]{xcolor}
\usepackage[utf8]{inputenc}
\usepackage{hyperref}
\usepackage{amsmath}
\usepackage{amssymb}
\usepackage{mathtools}
\usepackage{amsthm}
\usepackage[most]{tcolorbox}

\usepackage[capitalize,noabbrev]{cleveref}

\usepackage{multirow}
\usepackage{multicol}

\usepackage{amsmath,amsfonts,bm}

\def\eqref#1{equation~\ref{#1}}

\def\1{\bm{1}}

\DeclareMathAlphabet{\mathsfit}{\encodingdefault}{\sfdefault}{m}{sl}
\SetMathAlphabet{\mathsfit}{bold}{\encodingdefault}{\sfdefault}{bx}{n}

\usepackage{hyperref}
\usepackage{url}

\title{Gradient Atoms:\\Unsupervised Discovery, Attribution and Steering of Model Behaviors via Sparse Decomposition of Training Gradients
}

\author{J Rosser \\
FLAIR, University of Oxford \\
\texttt{jrosser@robots.ox.ac.uk}
}

\iclrfinalcopy 
\runningtitle{Gradient Atoms}
\begin{document}

\maketitle

\begin{abstract}
Training data attribution (TDA) methods ask which training documents are responsible for a model behavior. However, models often learn broad concepts shared across many examples.
Moreover, existing TDA methods are supervised---they require a predefined query behavior, then score every training document against it---making them both expensive and unable to surface behaviors the user did not think to ask about.
We present \textbf{Gradient Atoms}, an unsupervised method that decomposes per-document training gradients into sparse components (``atoms'') via dictionary learning in a preconditioned eigenspace. Each atom captures a shared update direction induced by a cluster of functionally similar documents, directly recovering the collective structure that per-document methods do not address.
Among 500 discovered atoms, the highest-coherence ones recover interpretable task-type behaviors---refusal, arithmetic, yes/no classification, trivia QA---without any behavioral labels. These atoms double as effective steering vectors: applying them as weight-space perturbations produces large, controllable shifts in model behavior (e.g., bulleted-list generation 33\% $\to$ 94\%; systematic refusal 50\% $\to$ 0\%). The method requires no query--document scoring stage, and scales independently of the number of query behaviors of interest. Code is available at \url{https://github.com/jrosseruk/gradient_atoms}.
\end{abstract}

\section{Introduction}
\label{sec:intro}

\begin{figure}[h!]
\centering
\includegraphics[width=\linewidth]{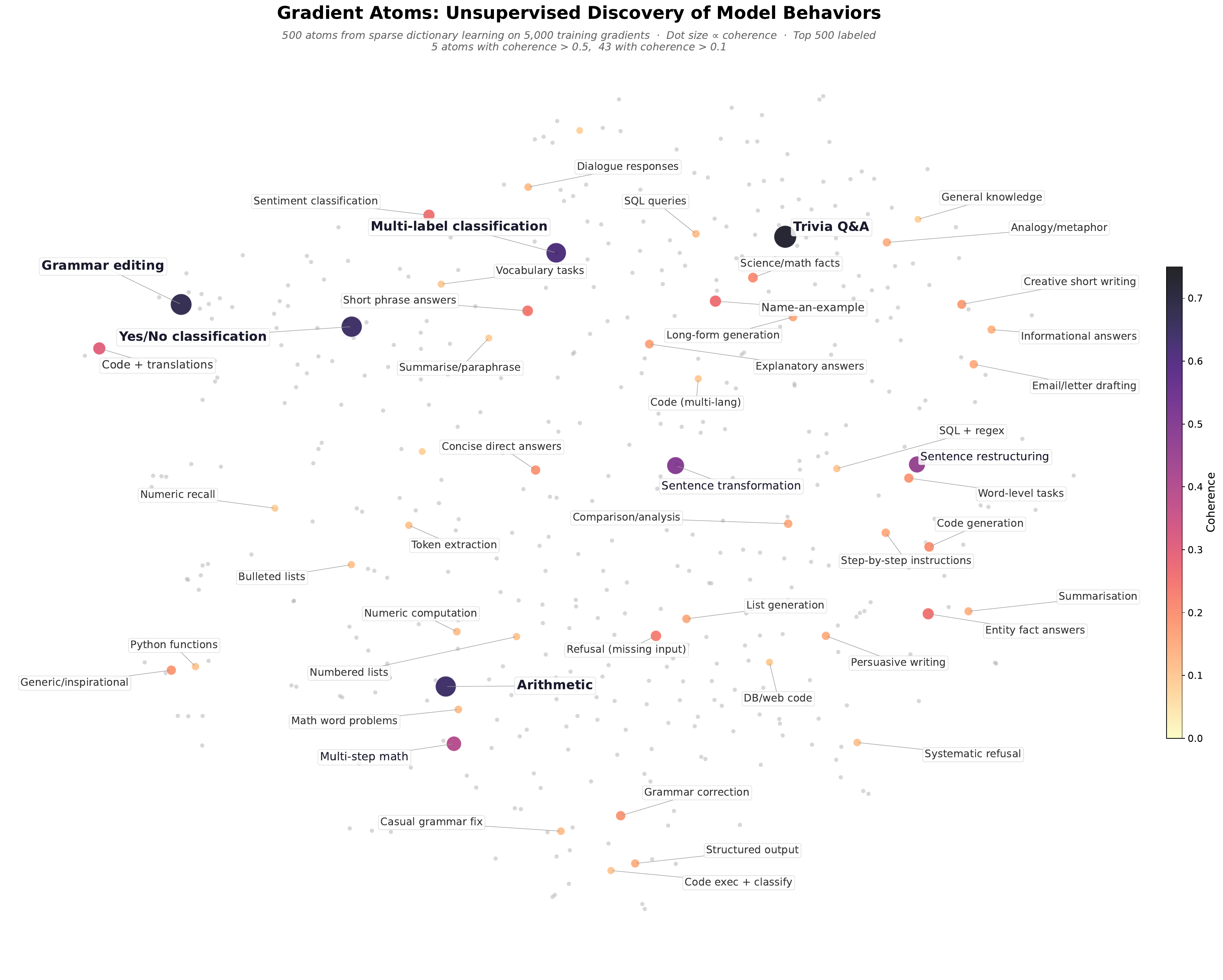}
\caption{Gradient atoms discovered via sparse dictionary learning over 5,000 training-document gradients. Each point is one atom; high-coherence atoms correspond to tightly defined task types.}
\label{fig:atoms_scatter}
\end{figure}

Training data attribution (TDA) attributes a model's behavior to the documents it was trained on ~\citep{koh2017understanding, grosse2023studying, source}. Per-document attribution has proven valuable for identifying influential examples, debugging predictions, and curating datasets.

However, per-document scoring captures only part of what training teaches. A model that learns to perform arithmetic during fine-tuning does so not because of any single arithmetic example, but because many arithmetic examples induce a shared update direction across weights. More broadly, \citet{ruis2024procedural} show that fine-tuning primarily instils \emph{procedural} capabilities---task-level strategies such as classification, editing, or code generation---that emerge from the collective gradient signal of functionally similar documents.

Standard TDA also faces a practical bottleneck: it is \textbf{supervised}, requiring the user to specify a query behavior before scoring documents against it. This demands (1) knowing what behaviors to look for in advance, and (2) an $O(N)$ scoring pass per query. For influence-function methods like EKFAC~\citep{grosse2023studying, source}, attributing $Q$ behaviors costs $O(Q \times N)$ query--document comparisons. For a comprehensive audit of learned behaviors, this is prohibitively expensive.

We address both gaps with a different question. Instead of asking ``which document caused this behavior?'', we ask: \textbf{what are the shared update directions that clusters of documents jointly induce?}

This is the idea behind \textbf{Gradient Atoms}. We extract per-document gradients, project them into a preconditioned eigenspace (via EKFAC), and apply sparse dictionary learning to decompose them into atoms. Each atom is a direction in weight space shared by a cluster of functionally similar documents. The method is:
\begin{itemize}
\item \textbf{Unsupervised}: no behavioral labels, measurement functions, or contrastive pairs are needed.
\item \textbf{Efficient}: one decomposition surfaces candidate behaviors simultaneously, with no per-query scoring stage.
\item \textbf{Actionable}: each atom can be unprojected into a full-parameter steering vector.
\end{itemize}

We validate this in two stages. First, we show that sparse dictionary learning on EKFAC-preconditioned gradients discovers 500 atoms, of which the highest-coherence atoms correspond to interpretable task types (\cref{tab:top_atoms}). Second, we show that these atoms---discovered without any behavioral labels---function as effective \textbf{steering vectors}: applying them as weight perturbations produces large, controllable shifts in five tested behaviors (\cref{fig:behavioral_steering}).

Our contributions are:
\begin{enumerate}
\item We show that per-document TDA provides an incomplete account of fine-tuning and propose a richer unit of analysis: shared update directions in gradient space.
\item We introduce Gradient Atoms, an unsupervised method that discovers candidate model behaviors from training gradients alone, without supervision or per-query scoring.
\item We demonstrate that discovered atoms function as effective steering vectors, producing large controllable shifts in model behavior without any behavioral labels.
\end{enumerate}

\section{Related Work}

Several concurrent works share our core insight that gradient similarity reflects functional similarity among training examples. GradientSpace~\citep{sridharan2025gradientspace} clusters LoRA gradients to identify ``latent skills'' but uses them to train specialised expert routers, not for interpretability or steering. Mode-Conditioning~\citep{wu2025modeconditioning} confirms that gradient clustering reliably recovers functional groupings (98.7\% F1), but applies this to test-time compute allocation. ELROND~\citep{skiers2026elrond} is philosophically closest: they decompose per-sample gradients into steerable directions via PCA and sparse autoencoders, but in diffusion models rather than LLMs.

Our contribution combines: (1) operating on training gradients rather than inference activations, (2) being fully unsupervised with no per-query scoring, and (3) the discovered atoms directly functioning as steering vectors for LLMs. Standard TDA methods~\citep{koh2017understanding, grosse2023studying, source} require a predefined query and $O(Q \times N)$ query--document comparisons; atoms complement this with unsupervised discovery. In the activation-space literature, SAEs~\citep{nanda2023progress} and gradient-informed variants~\citep{olmo2024gsae, shu2025gradsae} decompose activations at inference time; atoms decompose what the model \emph{learned} during training, across all layers simultaneously. Extended discussion is in \cref{app:related_work}.

\section{Method}

Consider a language model with parameters $\theta$ fine-tuned on a dataset of $N$ input--output pairs $\{x_1, \ldots, x_N\}$. Each document $x_i$ induces a gradient $g_i = \nabla_\theta \mathcal{L}_{\text{CE}}(\theta; x_i) \in \mathbb{R}^d$, where $d$ is the number of trainable parameters. The gradient $g_i$ is the direction the model's weights would move to improve on document $i$. Gradient Atoms decomposes these per-document gradients into sparse components; the highest-quality components isolate interpretable learned behaviors. The pipeline has five steps.

\subsection{Per-Document Gradient Extraction}

For each training document $x_i$, we compute the gradient of cross-entropy loss with respect to all trainable parameters, producing a gradient matrix $G \in \mathbb{R}^{N \times d}$. Documents that require similar computations tend to produce similar gradient vectors.

\subsection{EKFAC Projection and Preconditioning}

The raw gradient space is anisotropic---some directions have high curvature (small weight changes cause large loss changes) and others have low curvature. Without correction, any decomposition is dominated by high-curvature directions, drowning out semantic structure. We use the EKFAC eigendecomposition~\citep{grosse2023studying, source} of the approximate Fisher information matrix to correct for this. For each module $m$ with eigenvectors $Q_m$ and eigenvalues $\lambda_m$, we project each gradient into the top-$k$ eigenvectors and precondition:
\begin{equation}
\tilde{g}_i^{(m)} = Q_m^{(k)\top} \, g_i^{(m)}, \qquad
\hat{g}_i^{(m)} = \frac{\tilde{g}_i^{(m)}}{\sqrt{\lambda_m^{(k)} + \epsilon}}
\end{equation}
After concatenating across all $M$ modules, the projected gradient is $\hat{g}_i \in \mathbb{R}^{k_{\text{total}}}$ where $k_{\text{total}} = k \times M$. This makes the space approximately isotropic: a unit step in any direction corresponds to a roughly equal change in loss, encouraging atoms to capture functionally distinct directions rather than curvature artifacts.

\subsection{Sparse Dictionary Learning}

We normalize each projected gradient to unit norm (so atoms reflect direction, not magnitude) and apply sparse dictionary learning to decompose:
\begin{equation}
\hat{g}_i \approx \sum_{j=1}^{K} \alpha_{ij} \, d_j
\end{equation}
where $D = [d_1, \ldots, d_K] \in \mathbb{R}^{K \times k_{\text{total}}}$ are the \textbf{atoms} and $\alpha_{ij}$ are sparse coefficients---most are zero, so each document is explained by a few atoms. The sparsity penalty encourages each atom to capture a single pattern rather than blending multiple unrelated behaviors.

\subsection{Coherence Scoring}

For each atom $j$, we identify its activating documents---those with non-zero coefficient $\alpha_{ij}$---and compute a coherence score over the top-$n$ activating documents $\mathcal{S}_j$:
\begin{equation}
\text{coherence}(j) = \frac{1}{|\mathcal{S}_j|(|\mathcal{S}_j|-1)} \sum_{a \neq b \in \mathcal{S}_j} \cos(g_a, g_b)
\end{equation}
where $g_a, g_b$ are the \emph{raw} (unprojected, full $d$-dimensional) gradients. High coherence suggests the atom has found a shared computational motif in the original weight space, rather than an artifact of the projection.

\subsection{Unprojection to Steering Vectors}

Any atom can be converted back to a full parameter-space vector by reversing the projection:
\begin{equation}
v_j = \text{unproject}(d_j) \in \mathbb{R}^d
\end{equation}
This vector can be applied as a weight-space perturbation $\theta_{\text{new}} = \theta \pm \alpha \cdot v_j$, analogous to curvature-aware model editing~\citep{ikram2026crispedit}. The key difference is that $v_j$ was discovered unsupervised from the training data, rather than derived from a hand-crafted measurement function or contrastive pair.

\section{Experiments}

\subsection{Setup}

\textbf{Model.} We use \textbf{Gemma-3 4B IT} fine-tuned via LoRA (rank 8) on the \texttt{q\_proj} and \texttt{v\_proj} matrices across all 34 layers, yielding 2.2M trainable parameters across 136 LoRA modules.

\textbf{Dataset.} The model is fine-tuned on 5,000 instruction--response pairs sampled from a general-purpose SFT mixture covering arithmetic, grammar correction, classification, code generation, QA, creative writing, and other tasks. EKFAC factors (eigendecomposition of the approximate Fisher) are computed on the full training set.

\textbf{Gradient Atoms pipeline.} We extract per-document gradients for all 5,000 training examples, project via EKFAC into 6,800 dimensions (50 eigencomponents $\times$ 136 modules, a $328\times$ reduction), and run MiniBatchDictionaryLearning (scikit-learn) with $K = 500$ atoms and sparsity penalty $\alpha = 0.1$. Coherence is computed over the top-20 activating documents per atom using the raw 2.2M-dimensional gradients.

\subsection{Atom Discovery}
\label{sec:discovery}

From 500 atoms: 5 have coherence $> 0.5$, 43 have coherence $> 0.1$, and 457 have coherence $< 0.1$. \Cref{fig:atoms_scatter} shows the distribution. The top 5 atoms (coherence $>0.5$) are: short factual QA (0.725), grammar editing (0.672), yes/no classification (0.647), simple arithmetic (0.643), and multi-category classification (0.614). The full top~50 are listed in \cref{tab:top_atoms}.

\textbf{Atoms capture task types, not topics.} The decomposition clusters training data by \emph{how} the model responds (arithmetic, classification, editing, code) rather than \emph{what} it responds about (science, history, culture). This is consistent with the procedural-knowledge hypothesis of \citet{ruis2024procedural} discussed in \cref{sec:intro}, and confirms that gradient structure is organised around shared computational strategies.

\textbf{High coherence = stereotyped tasks.} The top-5 atoms are all formulaic task types whose activating documents have highly similar gradients, suggesting similar computational pathways.

\textbf{Multiple granularities.} Grammar correction appears three times (ranks 2, 17, 36) and code generation five times (ranks 16, 37, 40, 45, 46), at decreasing coherence. The dictionary finds sub-clusters that may reflect different sentence complexity levels or programming language families.

\textbf{Format atoms.} Bulleted lists (\#469) and numbered lists (\#299) are separate atoms, suggesting the model uses distinct weight pathways for these formatting patterns.

\textbf{Refusal is discoverable.} Two atoms (\#52, \#161) capture the model's tendency to reply ``Please provide the input'' when task instructions lack content---a behavior learned from training data that appears to be separable from other behaviors in gradient space.

\textbf{Effect of sparsity penalty.} \Cref{tab:sparsity} shows the effect of $\alpha$. At $\alpha = 0.01$, atoms are too dense (median ${\sim}2500$ docs each), blending unrelated patterns. At $\alpha = 0.1$, atoms are selective (${\sim}100$ docs each). At $\alpha = 1.0$, the penalty overwhelms reconstruction and all coefficients are zero.

\begin{table}[h!]
\centering
\caption{Effect of sparsity penalty $\alpha$ on atom quality.}
\label{tab:sparsity}
\begin{tabular}{rccc}
\toprule
$\alpha$ & Docs per atom (median) & Atoms coh $> 0.5$ & Atoms coh $> 0.1$ \\
\midrule
0.01 & ${\sim}2500$ & 3 & ${\sim}20$ \\
0.1  & ${\sim}100$  & 5 & 43 \\
1.0  & 0            & 0 & 0 \\
\bottomrule
\end{tabular}
\end{table}

\subsection{Behavioral Steering}
\label{sec:steering}

A key test of whether gradient atoms capture genuine computational structure is whether they can \textbf{steer model behavior} when applied as weight-space perturbations. We select five atoms spanning a range of coherence scores and behavioral types, unproject each to a full LoRA parameter-space vector, and apply perturbations $\theta_{\text{new}} = \theta \pm \alpha \cdot v_j$ with $\alpha \in \{0.5, 1.0, 2.0, 5.0, 10.0\}$ in both directions. Since dictionary learning assigns atom signs arbitrarily, we test both. For each atom, we design 100 evaluation questions (${\sim}60$ that naturally invite the target behavior, ${\sim}40$ neutral controls) and measure behavior with a regex detector:
\begin{itemize}
\item \textbf{Yes/No} (\#415): first line starts with Yes/No/True/False
\item \textbf{Code} (\#64): response contains a fenced code block (\texttt{```})
\item \textbf{Refusal} (\#161): matches clarification-seeking patterns
\item \textbf{Bullets} (\#469): $\geq 2$ lines starting with \texttt{-}, \texttt{*}, or $\bullet$
\item \textbf{Numbered} (\#299): $\geq 2$ lines starting with \texttt{\textbackslash d+[.)]}
\end{itemize}

Each steered adapter is served via vLLM alongside the clean baseline, with all 11 variants (5 alphas $\times$ 2 signs + baseline) loaded simultaneously as LoRA modules.

\begin{figure}[h!]
\centering
\includegraphics[width=\linewidth]{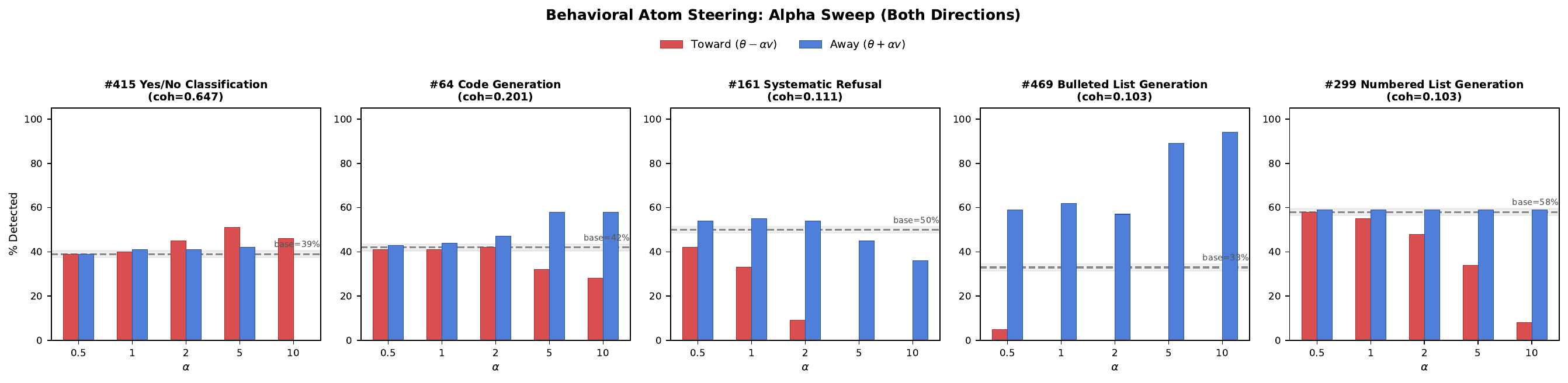}
\caption{Behavioral steering via unsupervised gradient atoms. Red bars show the ``toward'' direction ($\theta - \alpha v$), blue bars show ``away'' ($\theta + \alpha v$), and the dashed line marks the clean baseline. Four of five atoms produce large, monotonic steering effects in at least one direction.}
\label{fig:behavioral_steering}
\end{figure}

\begin{table}[h!]
\centering
\caption{Behavioral atom steering results. For each atom, we report the baseline detection rate and the best increase/decrease achieved across all alpha values and both directions. $\Delta$ is the change in percentage points from baseline.}
\label{tab:steering_results}
\small
\begin{tabular}{rlcccccc}
\toprule
Atom & Behavior & Coh. & Base & Best $\uparrow$ & $\Delta\!\uparrow$ & Best $\downarrow$ & $\Delta\!\downarrow$ \\
\midrule
\#415 & Yes/No Classification & 0.647 & 39\% & 51\% & $+$12pp & 0\% & $-$39pp \\
\#64  & Code Generation       & 0.201 & 42\% & 58\% & $+$16pp & 28\% & $-$14pp \\
\#161 & Systematic Refusal    & 0.111 & 50\% & 55\% & $+$5pp  & 0\%  & $-$50pp \\
\#469 & Bulleted Lists        & 0.103 & 33\% & 94\% & $+$61pp & 0\%  & $-$33pp \\
\#299 & Numbered Lists        & 0.103 & 58\% & 59\% & $+$1pp  & 8\%  & $-$50pp \\
\bottomrule
\end{tabular}
\end{table}

Results are shown in \cref{fig:behavioral_steering} and \cref{tab:steering_results}. All five atoms steer behavior in at least one direction, and four produce large effects ($>$14pp):

\begin{itemize}
\item \textbf{Bulleted lists (\#469)} is the strongest: steering increases bullet usage from 33\% to 94\% ($+$61pp) and suppresses it to 0\%. The effect is monotonic with alpha in both directions.

\item \textbf{Refusal (\#161)} is completely suppressible: from 50\% baseline to 0\% at $\alpha = 5$. The steered model responds ``Okay.'' to underspecified prompts instead of asking for clarification. The reverse direction modestly increases refusal ($+$5pp) and makes the model more verbose, suggesting this atom captures a terse/verbose dimension.

\item \textbf{Code generation (\#64)} increases from 42\% to 58\% ($+$16pp) or decreases to 28\% ($-$14pp). The sign is flipped relative to the Newton step convention, confirming that atom signs from dictionary learning are arbitrary.

\item \textbf{Yes/No (\#415)} shows moderate amplification ($+$12pp) but strong suppression ($-$39pp to 0\%). At high alpha, the model loses coherence, establishing the upper bound of useful perturbation.

\item \textbf{Numbered lists (\#299)} is asymmetric: easily suppressed (58\% $\to$ 8\%) but not amplifiable ($+$1pp), likely due to a ceiling effect.
\end{itemize}

\section{Discussion}

\textbf{Suppression appears easier than amplification.} All five atoms suppress their target behavior to near zero, but only two (bullets, code) achieve substantial amplification. One interpretation is that suppressing a behavior requires disrupting a single computational pathway, while amplifying requires strengthening it against many competing alternatives.

\textbf{Coherence does not necessarily predict steerability.} Atom \#469 (coherence 0.103) produces the largest steering effect ($+$61pp), while \#415 (coherence 0.647) gives only $+$12pp. Coherence measures gradient alignment among activating documents, but steerability also depends on how much the model's default behavior already saturates the target pathway.

\textbf{Limitations.}
Only 43 of 500 atoms have coherence $>0.1$, the majority are noise or capture overly broad mixtures. The instruction-following training data means atoms recover task types rather than fine-grained semantic preferences; more naturalistic data might yield different atoms. The 6,800-dim EKFAC projection discards information, and 5,000 documents may not cover rare behaviors. Our regex-based evaluation measures surface formatting rather than deeper behavioral changes.

\section{Conclusion}

We presented Gradient Atoms, an unsupervised method that discovers what a fine-tuning dataset teaches a model by decomposing training gradients into sparse components. The method addresses a gap in standard training data attribution: rather than scoring individual documents against a known behavior, it recovers the shared update directions that clusters of documents jointly induce---the collective structure through which procedural capabilities are acquired. The highest-coherence atoms recover interpretable task-type behaviors without supervision, and function as effective steering vectors, connecting unsupervised behavior discovery with controllable model editing.

Future directions include composing multiple atoms for simultaneous multi-behavior steering, scaling the dictionary to 1,000+ atoms, cross-model comparison to identify shared vs.\ adapter-specific behaviors, and developing principled methods for alpha selection.




\section*{Acknowledgements}

J Rosser is supported by the EPSRC centre for Doctoral Training in Autonomous and Intelligent Machines and Systems EP/Y035070/1. Special thanks to the London Initiative for Safe AI and Arcadia Impact for providing workspace.


\bibliography{iclr2026_conference}
\bibliographystyle{iclr2026_conference}

\appendix
\break
\section{Extended Related Work}
\label{app:related_work}

\textbf{Training data attribution.}
Influence functions~\citep{cook1982residuals, koh2017understanding} estimate the effect of removing a training example on predictions. \citet{grosse2023studying} scaled these to LLMs via EKFAC, and \citet{source} improved accuracy with approximate unrolled differentiation. All such methods are supervised---requiring a query behavior and $O(N)$ scoring per query. Gradient Atoms complements these approaches by discovering candidate behaviors without a predefined query.

\textbf{Gradient-based clustering.}
GradientSpace~\citep{sridharan2025gradientspace} clusters LoRA gradients via online SVD to identify ``latent skills,'' sharing our core insight that gradient similarity reflects functional similarity. However, they use clusters to train specialised LoRA experts with a router, not for interpretability or steering. Their SVD + k-means yields routing labels; our sparse dictionary learning produces individually steerable atoms. Mode-Conditioning~\citep{wu2025modeconditioning} independently confirms that gradient clustering recovers functional groupings (98.7\% F1), applying this to test-time compute allocation.

\textbf{Gradient decomposition in diffusion models.}
ELROND~\citep{skiers2026elrond} decomposes per-sample gradients via PCA and sparse autoencoders into steerable directions for visual attribute control---the diffusion-model analogue of our approach. Key differences: they decompose gradients \emph{within} a single prompt's realisations rather than across the full training set, and target visual attributes rather than LLM behaviors.

\textbf{Activation-space interpretability.}
SAEs decompose single-layer activations into monosemantic features~\citep{nanda2023progress}; gradient-informed variants (g-SAEs~\citep{olmo2024gsae}, GradSAE~\citep{shu2025gradsae}) use gradients to improve feature selection. Both still decompose activations---we decompose the gradients themselves, across all layers simultaneously. \citet{wang2026patterning} frame the theoretical dual of interpretability (behavior $\to$ training causes); Gradient Atoms provides a practical mechanism that additionally discovers behaviors without a query.

\textbf{Model editing and procedural knowledge.}
Prior steering methods require a known concept with measurement functions or contrastive pairs~\citep{ikram2026crispedit}; Gradient Atoms discovers steering directions unsupervised. Our finding that atoms capture task types rather than topics is consistent with \citet{ruis2024procedural}, who show procedural knowledge drives what models extract from training data.

\section{Computational Details}
\label{app:compute}

\begin{table}[h]
\centering
\caption{Computational cost of the Gradient Atoms pipeline.}
\label{tab:compute}
\begin{tabular}{lll}
\toprule
Step & Resources & Time \\
\midrule
Gradient extraction & $8\times$ A100 40GB & 170s \\
EKFAC projection & CPU, 16GB RAM & ${\sim}5$ min \\
Dictionary learning ($\alpha=0.1$) & CPU, 32GB RAM & ${\sim}15$ min \\
Coherence computation & CPU, 8GB RAM & ${\sim}5$ min \\
\midrule
\textbf{Total} & & $\mathbf{{\sim}25}$ \textbf{min} \\
\bottomrule
\end{tabular}
\end{table}

Model: Gemma-3 4B IT, LoRA rank 8 (\texttt{q\_proj} + \texttt{v\_proj}), 2.2M parameters, 136 modules across 34 layers. EKFAC factors computed separately on the full training set.

\section{Full Atom Table}
\label{app:atoms}

\begin{table}[h!]
\centering
\caption{Top 50 gradient atoms ranked by coherence score. Each atom was characterised by manual inspection of its top-20 activating documents.}
\label{tab:top_atoms}
\small
\begin{tabular}{rrrrl}
\toprule
Rank & Atom & Coherence & Active Docs & Description \\
\midrule
1 & \#348 & 0.725 & 139 & Short factual Q\&A---trivia with one-word/numeric answers \\
2 & \#328 & 0.672 & 110 & Grammar and sentence editing \\
3 & \#415 & 0.647 & 156 & Yes/No/True/False binary classification \\
4 & \#458 & 0.643 & 124 & Simple arithmetic \\
5 & \#498 & 0.614 & 176 & Multi-category classification and labeling \\
6 & \#358 & 0.499 & 88 & Sentence transformation (voice, tense, translation) \\
7 & \#2 & 0.463 & 206 & Sentence restructuring (questions, passive/active) \\
8 & \#451 & 0.395 & 182 & Multi-step arithmetic and unit conversions \\
9 & \#484 & 0.298 & 49 & Mixed technical (code + translations + set ops) \\
10 & \#319 & 0.262 & 180 & ``Name an example of X''---single-entity retrieval \\
11 & \#430 & 0.258 & 150 & Sentiment and text classification \\
12 & \#425 & 0.257 & 215 & Single-entity factual answers \\
13 & \#363 & 0.238 & 146 & Short phrase answers to open questions \\
14 & \#52 & 0.230 & 57 & ``Please provide the input''---refusal on missing input \\
15 & \#364 & 0.205 & 158 & Science and math fact answers \\
16 & \#64 & 0.201 & 25 & Code generation (Python, JS, C++, HTML) \\
17 & \#303 & 0.189 & 49 & Grammar correction on short sentences \\
18 & \#394 & 0.188 & 144 & Concise direct answers (mixed tasks) \\
19 & \#488 & 0.187 & 168 & Short inspirational/generic responses \\
20 & \#477 & 0.185 & 227 & Word-level tasks (synonyms, antonyms, rhymes) \\
21 & \#376 & 0.176 & 97 & Creative short-form writing \\
22 & \#136 & 0.165 & 161 & Multi-sentence explanatory answers \\
23 & \#66 & 0.154 & 45 & Long-form generation (essays, paragraphs) \\
24 & \#457 & 0.152 & 83 & Comparison and analysis tasks \\
25 & \#256 & 0.152 & 50 & Step-by-step instructions and how-to guides \\
26 & \#265 & 0.151 & 118 & List generation (brainstorming, idea lists) \\
27 & \#224 & 0.149 & 31 & Email and letter drafting \\
28 & \#446 & 0.146 & 86 & Persuasive/argumentative writing \\
29 & \#294 & 0.142 & 50 & Data extraction and structured output \\
30 & \#419 & 0.142 & 78 & Analogy and metaphor reasoning \\
31 & \#359 & 0.137 & 211 & Neutral informational answers \\
32 & \#306 & 0.136 & 118 & Summarisation \\
33 & \#181 & 0.119 & 87 & Dialogue and conversational responses \\
34 & \#72 & 0.118 & 37 & Math word problems \\
35 & \#445 & 0.117 & 69 & Numeric computation (GCF, LCM, time) \\
36 & \#465 & 0.116 & 70 & Grammar correction on casual sentences \\
37 & \#231 & 0.115 & 21 & SQL queries and structured code \\
38 & \#161 & 0.111 & 47 & Systematic refusal on unclear input \\
39 & \#325 & 0.106 & 143 & Single-word/token extraction from input \\
40 & \#61 & 0.105 & 21 & Python utility function implementations \\
41 & \#469 & 0.103 & 143 & Bulleted list generation \\
42 & \#299 & 0.103 & 46 & Numbered list generation \\
43 & \#67 & 0.102 & 9 & SQL + regex + technical expressions \\
44 & \#180 & 0.100 & 52 & Mixed code execution and classification \\
45 & \#381 & 0.097 & 79 & Code generation (broad, multi-language) \\
46 & \#428 & 0.096 & 81 & Database/web code (SQL, HTML, CSS, APIs) \\
47 & \#48 & 0.095 & 56 & Single-word vocabulary tasks (fill-in-blank, plurals) \\
48 & \#475 & 0.088 & 83 & Summarisation and paraphrasing \\
49 & \#233 & 0.087 & 46 & Numeric/factual recall with approximation \\
50 & \#172 & 0.084 & 129 & General knowledge Q\&A \\
\bottomrule
\end{tabular}
\end{table}

\end{document}